%% file: main.tex
\documentclass[conference]{IEEEtran}
\IEEEoverridecommandlockouts

\usepackage{booktabs}
\usepackage{cite}
\usepackage{amsmath,amssymb,amsfonts}
\usepackage{algorithmic}
\usepackage{graphicx}
\usepackage{textcomp}
\usepackage{xcolor}

\usepackage{marvosym}
\usepackage{tikz}
\usepackage{subfigure}
\usepackage{pgfplots}
\usepackage{multirow}
\usepackage{rotating}
\usepackage{makecell}
\usepackage{hyperref}
\hypersetup{
     colorlinks=true,
     linkcolor=blue,
     filecolor=magenta,
     citecolor = black,      
     urlcolor=cyan,
     }
\usepackage{mathrsfs}
\usetikzlibrary{arrows}
\usetikzlibrary{positioning}

\pgfplotsset{compat=1.17}

\def\BibTeX{{\rm B\kern-.05em{\sc i\kern-.025em b}\kern-.08em
    T\kern-.1667em\lower.7ex\hbox{E}\kern-.125emX}}

\title{
Improving Model-Based Reinforcement Learning with Internal State Representations through Self-Supervision
}

\begin{document}

\author{\IEEEauthorblockN{Julien Scholz, Cornelius Weber, Muhammad Burhan Hafez and Stefan Wermter }
\IEEEauthorblockA{\textit{Knowledge Technology, Department of Informatics, University of Hamburg, Germany}\\
julienscholzo@hotmail.de, \{weber, hafez, wermter\}@informatik.uni-hamburg.de}
}

\maketitle

\input{abstract.tex}

\input{introduction/introduction.tex}

\input{introduction/background.tex}
\input{proposed_methods/proposed_methods.tex}

\input{experiments/experiments.tex}

\input{conclusion/conclusion.tex}

\section*{Acknowledgment}
The authors gratefully acknowledge support from the German Research Foundation DFG under project CML (TRR 169).

\bibliographystyle{IEEEtran}
\bibliography{main}

\end{document}

%% file: abstract.tex
\begin{abstract}
    Using a model of the environment, reinforcement learning agents can plan their future moves and achieve super-human performance in board games like Chess, Shogi, and Go, while remaining relatively sample-efficient. As demonstrated by the MuZero Algorithm, the environment model can even be learned dynamically, generalizing the agent to many more tasks while at the same time achieving state-of-the-art performance. Notably, MuZero uses internal state representations derived from real environment states for its predictions. In this paper, we bind the model's predicted internal state representation to the environment state via two additional terms: a reconstruction model loss and a simpler consistency loss, both of which work independently and unsupervised, acting as constraints to stabilize the learning process.
    Our experiments show that this new integration of reconstruction model loss and simpler consistency loss provide a significant performance increase in OpenAI Gym environments.
    Our modifications also enable self-supervised pretraining for MuZero, so the algorithm can learn about environment dynamics before a goal is made available.
\end{abstract}

%% file: introduction/introduction.tex
\section{Introduction}
Reinforcement learning algorithms are classified into two groups. Model-free algorithms can be viewed as behaving instinctively, while model-based algorithms can plan their next moves ahead of time. The latter are arguably more human-like and have the added advantage of being comparatively sample-efficient.

The \textit{MuZero Algorithm} \cite{muzero} outperforms its competitors in a variety of tasks while simultaneously being very sample-efficient. Unlike other recent model-based reinforcement learning approaches that perform gradient-based planning \cite{srinivas2018universal,ebert2018visual,Boots-RSS-19,hafez2019curious,hafner2019learning}, MuZero does not assume a continuous and differentiable action space. However, we realized that MuZero performed relatively poorly on our experiments in which we trained a robot to grasp objects using a low-dimensional input space and was comparatively slow to operate. Contrary to our expectations, older, model-free algorithms such as \textit{A3C} \cite{a3c} delivered significantly better results while being more lightweight and comparatively easy to implement. Even in very basic tasks, our custom implementation as well as one provided by other researchers, \textit{muzero-general} \cite{muzero-general}, produced unsatisfactory outcomes.

Since these results did not match those that MuZero achieved in complex board games \cite{muzero}, we are led to believe that MuZero may require extensive training time and expensive hardware to reach its full potential. We question some of the design decisions made for MuZero, namely, how unconstrained its learning process is. After explaining all necessary fundamentals, we propose two individual augmentations to the algorithm that work fully unsupervised in an attempt to improve its performance and make it more reliable. Afterward, we evaluate our proposals by benchmarking the regular MuZero algorithm against the newly augmented creation.

%% file: introduction/background.tex
\section{Background}
\input{introduction/muzero.tex}
\input{introduction/monte_carlo_tree_search.tex}

%% file: introduction/muzero.tex
\newcommand{\policy}{\text{\textbf{p}}}
\newcommand{\svalue}{\nu}
\newcommand{\given}{\,\middle|\,}

\subsection{MuZero Algorithm}
The MuZero algorithm \cite{muzero} is a model-based reinforcement learning algorithm that builds upon the success of its predecessor, \textit{AlphaZero} \cite{alphazero}, which itself is a generalization of the successful \textit{AlphaGo} \cite{silver2016mastering} and \textit{AlphaGo Zero} \cite{silver2017mastering} algorithms to a broader range of challenging board and Atari games. Similar to other model-based algorithms, MuZero can predict the behavior of its environment to plan and choose the most promising action at each timestep to achieve its goal. In contrast to AlphaZero, it does this using a learned model. As such, it can be applied to environments for which the rules are not known ahead of time.

MuZero uses an \textit{internal} (or \textit{embedded}) state representation that is deduced from the environment observation but is not required to have any semantic meaning beyond containing sufficient information to predict rewards and values. Accordingly, it may be infeasible to reconstruct observations from internal states. This gives MuZero an advantage over traditional model-based algorithms that need to be able to predict future observations (e.g. noisy visual input).

There are three distinct functions (e.g. neural networks) that are used harmoniously for planning. Namely, there is a \textit{representation} function $h_\theta$, a \textit{prediction} function $f_\theta$, as well as a \textit{dynamics} function $g_\theta$, each being parameterized using parameters $\theta$ to allow for adjustment through a training process. The complete model is called $\mu_\theta$. We will now explain each of the three functions in more detail.
\begin{itemize}
    \item The representation function $h_\theta$ maps real observations to internal state representations. For a sequence of recorded observations $o_1, \dots, o_t$ at timestep $t$, an embedded representation $s^0 = h(o_1, \dots, o_t)$ may be produced. As previously mentioned, $s^0$ has no semantic meaning, and typically contains significantly less information than $o_1, \dots, o_t$. Thus, the function $h_\theta$ is tasked with eliminating unnecessary details from observations, for example by extracting object coordinates and other attributes from images.

    \item The dynamics function $g_\theta$ tries to mimic the environment by advancing an internal state $s^{k-1}$ at a hypothetical timestep $k-1$ based on a chosen action $a^k$ to predict $r^k, s^k = g_\theta\left(s^{k-1}, a^k\right)$, where $r^k$ is an estimate of the real reward $u_{t+k}$ and $s^k$ is the internal state at timestep $k$. This function can be applied recursively, and acts as the simulating part of the algorithm, estimating what may happen when taking a sequence of actions $a^1, ..., a^k$ in a state $s^0$.

    \item The prediction function $f_\theta$ can be compared to an actor-critic architecture having both a policy and a value output. For any internal state $s^k$, there shall be a mapping $\policy^k, v^k = f_\theta(s^k)$, where policy $\policy^k$ represents the probabilities with which the agent would perform actions and value $v^k$ is the expected return in state $s^k$. Whereas the value is very useful to bootstrap future rewards after the final step of planning by the dynamics function, the policy of the prediction function may be counterintuitive, keeping in mind that the agent should derive its policy from the considered plans. We will explore the uses of $\policy^k$ further on.
\end{itemize}

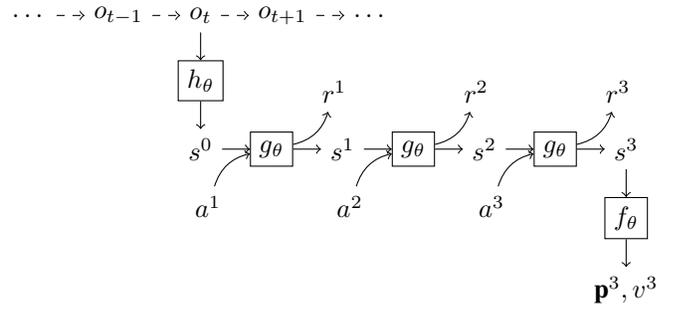
\begin{figure}[t]
    \centering
    \begin{tikzpicture}[node distance=0.37]
        \node (ot) {$o_t$};
        \node [left = of ot] (otm1) {$o_{t-1}$};
        \node [right = of ot] (otp1) {$o_{t+1}$};
        \node [left = of otm1] (otm2) {$\dots$};
        \node [right = of otp1] (otp2) {$\dots$};
        \draw [->, dashed] (otm2) -- (otm1);
        \draw [->, dashed] (otm1) -- (ot);
        \draw [->, dashed] (ot) -- (otp1);
        \draw [->, dashed] (otp1) -- (otp2);

        \node [below = of ot, draw] (h) {$h_\theta$};
        \node [below = of h] (s0) {$s^0$};
        \draw [->] (ot) -- (h);
        \draw [->] (h) -- (s0);

        \node [right = of s0, draw] (g0) {$g_\theta$};
        \node [right = of g0] (s1) {$s^1$};
        \node [below left = of g0] (a1) {$a^1$};
        \node [above right = of g0] (r1) {$r^1$};
        \draw [->] (s0) -- (g0);
        \draw [->] (a1) edge [bend left] (g0);
        \draw [->] (g0) -- (s1);
        \draw [->] (g0) edge [bend right] (r1);

        \node [right = of s1, draw] (g1) {$g_\theta$};
        \node [right = of g1] (s2) {$s^2$};
        \node [below left = of g1] (a2) {$a^2$};
        \node [above right = of g1] (r2) {$r^2$};
        \draw [->] (s1) -- (g1);
        \draw [->] (a2) edge [bend left] (g1);
        \draw [->] (g1) -- (s2);
        \draw [->] (g1) edge [bend right] (r2);

        \node [right = of s2, draw] (g2) {$g_\theta$};
        \node [right = of g2] (s3) {$s^3$};
        \node [below left = of g2] (a3) {$a^3$};
        \node [above right = of g2] (r3) {$r^3$};
        \draw [->] (s2) -- (g2);
        \draw [->] (a3) edge [bend left] (g2);
        \draw [->] (g2) -- (s3);
        \draw [->] (g2) edge [bend right] (r3);

        \node [below = of s3, draw] (f) {$f_\theta$};
        \node [below = of f] (pv) {$\policy^3, v^3$};
        \draw [->] (s3) -- (f);
        \draw [->] (f) -- (pv);
    \end{tikzpicture}
    \caption{Testing a single action sequence made up of three actions $a^1$, $a^2$, and $a^3$ to plan based on observation $o_t$ using all three MuZero functions.}
    \label{fig:muzero_basic_policy}
\end{figure}

Given parameters $\theta$, we can now decide on an action policy $\pi$ for each observation $o_t$, which we will call the \textit{search policy}, that uses $h_\theta$, $g_\theta$ and $f_\theta$ to search through different action sequences and find an optimal plan. As an example, in a small action space, we can iterate through all action sequences $a_1, ..., a_n$ of a fixed length $n$, and apply each function in the order visualized in Fig. \ref{fig:muzero_basic_policy}. By discounting the reward and value outputs with a discount factor $\gamma \in [0, 1]$, we receive an estimate for the return when first performing the action sequence and subsequently following $\pi$:
\begin{multline}
    \mathbb{E}_\pi\left[
        u_{t+1} + \gamma u_{t+2} + \dots \given o_t, a_1, \dots, a_n
    \right] \\ 
    = \sum_{k=1}^n \gamma^{k-1} r^k + \gamma^n v^n
\end{multline}
Given the goal of an agent to maximize the return, we may now simply choose the first action of the action sequence with the highest return estimate. Alternatively, a less promising action can be selected to encourage the exploration of new behavior. At timestep $t$, we call the return estimate for the chosen action produced by our search $\svalue_t$.

Training occurs on trajectories previously observed by the MuZero agent. For each timestep $t$ in the trajectory, we unroll the dynamics function $K$ steps through time and adjust $\theta$ such that the predictions further match what was already observed in the trajectory. Each reward output $r^k_t$ of $g_\theta$ is trained on the real reward $u_{t+k}$ through a loss term $l^r$. We also apply the prediction function $f_\theta$ to each of the internal states $s^0_t, ..., s^K_t$, giving us policy predictions $\policy^0_t, ..., \policy^K_t$ and value predictions $v^0_t, ..., v^K_t$. Each policy prediction $\policy^k_t$ is trained on the stored search policy $\pi_{t+k}$ with a policy loss $l^p$. This makes $\policy$ an estimate (that is faster to compute) of what our search policy $\pi$ might be, meaning it can be used as a heuristic. For our value estimates $\svalue$, we first calculate $n$-step bootstrapped returns $z_t = u_{t+1} + \gamma u_{t+2} + ... + \gamma^{n-1} u_{t+n} + \gamma^n\svalue_{t+n}$. With another loss $l^v$, the target $z_{t+k}$ is set for each value output $v^k_t$. The three previously mentioned output targets as well as an additional \textit{L2 regularization} term $c||\theta||^2$ \cite{l2-regularization} form the complete loss function:
\begin{multline}
    l_t(\theta) = \sum^K_{k=0} \Big(l^r(u_{t+k}, r_t^k) + l^v(z_{t+k}, v_t^k) \\
    + l^p(\pi_{t+k}, \policy_t^k) + c||\theta||^2\Big)
\end{multline}

The aforementioned brute-force search policy is highly unoptimized. For one, we have not described a method of caching and reusing internal states and prediction function outputs so as to reduce the computational footprint. Furthermore, iterating through all possible actions is very slow for a high number of actions or longer action sequences and impossible for continuous action spaces. Ideally, we would focus our planning efforts on only those actions that are promising from the beginning and spend less time incorporating clearly unfavorable behavior. AlphaZero and MuZero employ a variation of \textit{Monte-Carlo tree search} (\textit{MCTS}) \cite{mcts} (see Section \ref{sec_MCTS}).

MuZero matches the state-of-the-art results of AlphaZero in the board games Chess and \textit{Shogi}, despite not having access to a perfect environment model. It even exceeded AlphaZero's unprecedented rating in the board game \textit{Go}, while using less computation per node, suggesting that it caches useful information in its internal states to gain a deeper understanding of the environment with each application of the dynamics function. Furthermore, MuZero outperforms humans and previous state-of-the-art agents at the Atari game benchmark, demonstrating its ability to solve tasks with long time horizons that are not turn-based.

%% file: introduction/monte_carlo_tree_search.tex
\newcommand{\argmax}{\operatornamewithlimits{argmax}}

\subsection{Monte-Carlo Tree Search}
\label{sec_MCTS}
We will now explain the tree search algorithm used in MuZero's planning processes, a variant of the Monte-Carlo tree search, in detail \cite{alphazero, muzero}. The tree, which is constructed over the course of the algorithm, consists of states that make up the nodes, and actions represented by edges. Each path in the tree can be viewed as a trajectory. Our goal is to find the most promising action, that is, the action which yields the highest expected return starting from the root state.

Let $S(s, a)$ denote the state we reach when following action $a$ in state $s$. For each edge, we keep additional data:
\begin{itemize}
    \item $N(s, a)$ shall store the number of times we have visited action $a$ in state $s$ during the search.
    \item $Q(s, a)$, similar to the Q-table in Q-learning, represents the action-value of action $a$ in state $s$.
    \item $P(s, a) \in [0, 1]$ is an action probability with $\sum_{a \in \mathscr{A}(s)} P(s, a) = 1$. In other words, $P$ defines a probability distribution across all available actions for each state $s$, i.e. a policy. We will see that these policies are taken from the policy output of the prediction function.
    \item $R(s, a)$ is the expected reward when taking action $a$ in state $s$. Again, these values will be taken directly from the model's outputs.
\end{itemize}

At the start of the algorithm, the tree is created with an initial state $s^0$ which, in the case of MuZero, can be derived through the use of the representation function on an environment state. The search is then divided into three stages that are repeated for a number of \textit{simulations}.
\begin{enumerate}
    \item In the \textit{selection} stage, we want to find the part of the tree that is most useful to be expanded next. We want to balance between further advancing already promising trajectories, and those that have not been explored sufficiently as they seem unfavorable.

    We traverse the tree, starting from the root node $s^0$, for $k=1 \dots l$ steps, until we reach the currently uninitialized state $s^l$ which shall become our new leaf state. At each step, we follow the edge (or action) that maximizes an upper confidence bound called pUCT\footnote{polynomical Upper Confidence Trees}:
    \begin{multline}
        a^k = \argmax_a \Bigg[
            Q(s, a) +
            P(s, a) \cdot
            \frac{\sqrt{\sum_b N(s,b)}}{1 + N(s, a)} \\
            \cdot \left(
                c_1 + \log \left(
                    \frac{\sum_b N(s, b) + c_2 + 1}{c_2}
                \right)
            \right)
        \Bigg]
    \end{multline}
    The $Q(s, a)$ term prioritizes actions leading to states with a higher value, whereas $P(s, a)$ can be thought of as a heuristic for promising states provided by the model. To encourage more exploration of the state space, these terms are balanced through the constants $c_1$ and $c_2$ with the visit counts of the respective edge. An edge that has been visited less frequently therefore receives a higher pUCT value.

    For each step $k < l$ of the traversal, we take note of $a^k$, $s^k = S \left(s^{k-1}, a^k \right)$, and $r^k = R \left(s^{k-1}, a^k \right)$. The entries for $S \left(s^{l-1}, a^l \right)$ and $R \left(s^{l-1}, a^l \right)$ are yet to be initialized.

    \item In the \textit{expansion} stage, we attach a new state $s^l$ to the tree. To determine this state, we use the MuZero algorithm's dynamics function $r^l, s^l = g_\theta \left(s^{l-1}, a^l\right)$, advancing the trajectory by a single step, and then store $S \left( s^{l-1}, a^l\right) = s^l$ and $R \left(s^{l-1}, a^l \right) = r^l$. Similarly, we compute $\policy^l, v^l = f_\theta \left(s^l\right)$ with the help of the prediction function. For each available subsequent action $a$ in state $s^l$, we store
    $N \left(s^l, a \right) = 0$,
    $Q \left(s^l, a \right) = 0$, and
    $P \left(s^l, a \right) = \policy^l(a)$.
    This completes the second stage of the simulation.

    \item For the final \textit{backup} stage, we update the $Q$ and $N$ values for all edges along our trajectory in reverse order. First, for $k = l \dots 0$, we create bootstrapped return estimates
    \begin{equation}
        G^k = \sum_{\tau=0}^{l - 1 - k} \gamma^\tau r_{k+1+\tau} + \gamma^{l - k} v^l
    \end{equation}
    for each state in our trajectory. We update the action-values associated with each edge on our trajectory with
    \begin{equation}
        Q \left(s^{k-1}, a^k\right) \leftarrow \frac{
            N \left(s^{k-1}, a^k \right) \cdot Q \left(s^{k-1}, a^k \right) + G^k
        }{
            N \left(s^{k-1}, a^k \right) + 1
        },
    \end{equation}
    which simply creates a cumulative moving average of the expected returns across simulations. Finally, we update the visit counts of all edges in our path:
    \begin{equation}
        N \left(s^{k-1}, a^k \right) \leftarrow  N \left(s^{k-1}, a^k \right) + 1
    \end{equation}
    \vspace{0.2cm}
\end{enumerate}

This completes the three stages of the Monte-Carlo tree search algorithm. However, there is an issue with our pUCT formula. The $P(s, a)$ term should never leave the interval $[0, 1]$, whereas $Q(s, a)$ is theoretically unbounded, and depends on the magnitude of environment rewards. This makes the two terms difficult to balance. Intuitively, we are adding up unrelated units of measurement. A simple solution is to divide $Q(s, a)$ by the maximum reward that can be observed in the environment, as a means of normalizing it. Unfortunately, the maximum reward may not be known, and adding prior knowledge for each environment would make MuZero less of a general-purpose algorithm. Instead, we normalize our Q-values dynamically through min-max normalization with other Q-values in the current search tree T:
\begin{equation}
    \overline{Q} \left(s^{k-1}, a^k\right) = \frac{
        Q \left(s^{k-1}, a^k\right) - \min_{s, a \in T} Q(s, a)
    }{
        \max{s, a \in T} Q(s, a) - \min_{s, a \in T} Q(s, a)
    }
\end{equation}
In our pUCT formula, we may simply replace $Q(s, a)$ with $\overline{Q}(s, a)$.

After the tree has been fully constructed, we may define a policy for the root state as
\begin{equation}
    p_a = \frac{N(a)^{1/T}}{\sum_b N(b)^{1/T}},
\end{equation}
where $p_a$ is the probability of taking action $a$ in state $s^0$, and $T$ is a temperature parameter further balancing between exploration and exploitation. The search value shall be computed from all action-values $Q(s^0, a)$ based on this policy.

%% file: proposed_methods/proposed_methods.tex
\section{Proposed Methods}
In this section, we propose two changes to the MuZero algorithm that may improve its performance. These changes are based on the idea that MuZero is very unconstrained in its embedded state representations, which, while making the algorithm very flexible, may harm the learning process.

Our changes extend MuZero by individually weighable loss terms, meaning they can be viewed as a generalization of the regular MuZero algorithm and may be used either independently or combined.

\input{proposed_methods/reconstruction_function.tex}
\input{proposed_methods/consistency_loss_term.tex}

%% file: proposed_methods/reconstruction_function.tex
\newcommand{\reconstruction}{h^{-1}}

\subsection{Reconstruction Function}
\begin{figure}[t]
    \centering
    \begin{tikzpicture}[node distance=0.84]
        \node (ot) {$o_t$};
        \node [left = of ot] (otm1) {$\dots$};
        \node [right = of ot] (otp1) {$o_{t+1}$};
        \node [right = of otp1] (otp2) {$o_{t+2}$};
        \node [right = of otp2] (otp3) {$o_{t+3}$};
        \node [right = of otp3] (otp4) {$\dots$};
        \draw [->, dashed] (otm1) -- (ot);
        \draw [->, dashed] (ot) -- (otp1);
        \draw [->, dashed] (otp1) -- (otp2);
        \draw [->, dashed] (otp2) -- (otp3);
        \draw [->, dashed] (otp3) -- (otp4);

        \node [below = 1.75 of ot] (s0) {$s^0_t$};
        \node [below = 1.75 of otp1] (s1) {$s^1_t$};
        \node [below = 1.75 of otp2] (s2) {$s^2_t$};
        \node [below = 1.75 of otp3] (s3) {$s^3_t$};
        \draw [->] (ot) -- node [left] {$h_\theta$} (s0);
        \draw [->] (s0) -- node [below] {$g_\theta$} (s1);
        \draw [->] (s1) -- node [below] {$g_\theta$} (s2);
        \draw [->] (s2) -- node [below] {$g_\theta$} (s3);

        \node [below right = 0.1 of ot] (notot) {$\hat{o}^0_t$};
        \node [below right = 0.1 of otp1] (nototp1) {$\hat{o}^1_t$};
        \node [below right = 0.1 of otp2] (nototp2) {$\hat{o}^2_t$};
        \node [below right = 0.1 of otp3] (nototp3) {$\hat{o}^3_t$};
        \node [below right = -0.2 of ot] {\Lightning};
        \node [below right = -0.2 of otp1] {\Lightning};
        \node [below right = -0.2 of otp2] {\Lightning};
        \node [below right = -0.2 of otp3] {\Lightning};

        \draw [->] (s0) edge [bend right] node [right] {$\reconstruction_\theta$} (notot);
        \draw [->] (s1) edge [bend right] node [right] {$\reconstruction_\theta$} (nototp1);
        \draw [->] (s2) edge [bend right] node [right] {$\reconstruction_\theta$} (nototp2);
        \draw [->] (s3) edge [bend right] node [right] {$\reconstruction_\theta$} (nototp3);
    \end{tikzpicture}
    \caption{The new reconstruction function $\reconstruction_\theta$ being used to predict future observations $o_{t+k}$ from internal states $s^k$ with the help of the representation function $h_\theta$ as well as the dynamics function $g_\theta$.}
    \label{fig:reconstruction_function}
\end{figure}
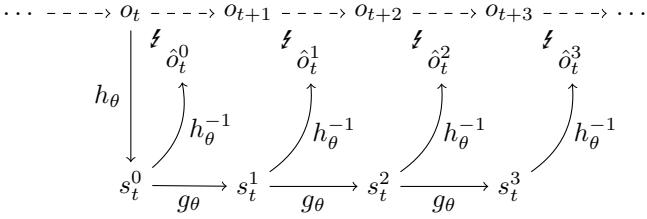

For our first proposed change, we introduce an additional $\theta$-parameterized function, which we call the \textit{reconstruction} function $\reconstruction_\theta$. Whereas the representation function $h_\theta$ maps observations to internal states, the reconstruction function shall perform the inverse operation, that is, mapping internal states to real observations, thereby performing a generative task. While optimizing reconstruction as an auxiliary task to learn state representations in reinforcement learning has been investigated \cite{lange2010deep,lange2012autonomous,kulkarni2016deep,de2018integrating,nair2020contextual}, reconstructing observations from states sampled from a learned dynamics model has not been considered. Notably, since we are using function approximation, reconstructed observations are unlikely to be perfectly accurate. Moreover, in the probable case that the embedded states store less information than the real observations, it becomes theoretically impossible for $\reconstruction_\theta$ to restore what has been discarded by the representation function. We call $\hat{o}^k_t = \reconstruction_\theta(s^k_t)$ the reconstructed observation for the embedded state $s^k_t$, meaning it is an estimate of the real observation $o_{t+k}$ (see Fig. \ref{fig:reconstruction_function}), since the dynamics function internally reflects the progression of the environment.

The reconstruction function is trained with the help of another loss term $l^g$ added to the default MuZero loss equation
\begin{multline}
    l_t(\theta) = \sum^K_{k=0} \big(l^r(u_{t+k}, r_t^k) + l^v(z_{t+k}, v_t^k) \\
    + l^p(\pi_{t+k}, \policy_t^k) + l^g(o_{t+k}, \hat{o}^k_t) + c||\theta||^2\big).
\end{multline}
This loss term shall be smaller the better the model becomes at estimating observations, for example, by determining the mean squared error between the real and the reconstructed observation. Notably, the error gradients must be propagated further through the dynamics function $g_\theta$, and eventually the representation function $h_\theta$. This means $h_\theta$ and $g_\theta$ are incentivized to maintain information that is useful for observation reconstruction.

Note that, so far, we have not specified any use cases for these reconstructed observations. We do not incorporate observation reconstruction in MuZero's planning process, and, in fact, the reconstruction function $\reconstruction_\theta$ may be discarded once the training process is complete. We are only concerned with the effects of the gradients for $l^g$ on the representation function $h_\theta$ and dynamics function $g_\theta$. To understand why, consider a MuZero agent navigating an environment with sparse rewards. It will observe many state transitions based on its actions that could reveal the potentially complex inner workings of the environment. However, it will skip out on gaining any insights unless it is actually given rewards, as the model's only goal is reward and value prediction. Even worse, the agent may be subject to \textit{catastrophic forgetting} of what has already been learned, as it is only being trained with a reward target of $0$ and small value targets. The reconstruction loss term $l^g$ shall counteract these issues by propagating its error gradients such that the representation function and dynamics function are forced, so to speak, to comprehend the environment beyond the rewards it supplies. Reconstructed observations are not meant to be accurate and their error gradients must not overpower the reward and value prediction. Instead, they should act merely as a guide to stabilize and accelerate learning.

An additional benefit is the ability to pretrain an agent in a self-supervised fashion. That is, the agent can explore an environment without being given any rewards (or any goal) in order to learn about its mechanics and develop a world model. This model can then be specialized to different goals within the same environment. The process is comparable to a child discovering a new task in an environment it is already familiar with, giving it an advantage by not having to learn from scratch.

%% file: proposed_methods/consistency_loss_term.tex
\subsection{Consistency Loss Term}
We additionally propose a simple loss term for MuZero's loss equation, which we call the \textit{consistency} loss and that does not require another function to be introduced into the system. The name originates from the possible inconsistencies in embedded state representations after each application of the dynamics function. MuZero is completely unconstrained in choosing a different internal state representation for each simulation step $k$.

\begin{figure}[t]
    \centering
    \begin{tikzpicture}[node distance=0.84]
        \node (ot) {$o_t$};
        \node [left = of ot] (otm1) {$\dots$};
        \node [right = of ot] (otp1) {$o_{t+1}$};
        \node [right = of otp1] (otp2) {$o_{t+2}$};
        \node [right = of otp2] (otp3) {$o_{t+3}$};
        \node [right = of otp3] (otp4) {$\dots$};
        \draw [->, dashed] (otm1) -- (ot);
        \draw [->, dashed] (ot) -- (otp1);
        \draw [->, dashed] (otp1) -- (otp2);
        \draw [->, dashed] (otp2) -- (otp3);
        \draw [->, dashed] (otp3) -- (otp4);

        \node [below = 1.75 of ot] (s0) {$s^0_t$};
        \node [below = 1.75 of otp1] (s1) {$s^1_t$};
        \node [below = 1.75 of otp2] (s2) {$s^2_t$};
        \node [below = 1.75 of otp3] (s3) {$s^3_t$};

        \draw [->] (ot) -- node [left] {$h_\theta$} (s0);
        \draw [->] (s0) -- node [below] {$g_\theta$} (s1);
        \draw [->] (s1) -- node [below] {$g_\theta$} (s2);
        \draw [->] (s2) -- node [below] {$g_\theta$} (s3);

        \node [above right = 0.1 of s1] (nots1) {$s^0_{t+1}$};
        \node [above right = -0.25 of s1] {\Lightning};
        \node [above right = 0.1 of s2] (nots2) {$s^0_{t+2}$};
        \node [above right = -0.25 of s2] {\Lightning};
        \node [above right = 0.1 of s3] (nots3) {$s^0_{t+3}$};
        \node [above right = -0.25 of s3] {\Lightning};

        \draw [->] (otp1) edge [bend right] node [left] {$h_\theta$} (nots1);
        \draw [->] (otp2) edge [bend right] node [left] {$h_\theta$} (nots2);
        \draw [->] (otp3) edge [bend right] node [left] {$h_\theta$} (nots3);
    \end{tikzpicture}
    \caption{Visualization of the discrepancy between state outputs of the dynamics function $g_\theta$ and representation function $h_\theta$.}
    \label{fig:consistency_loss}
\end{figure}
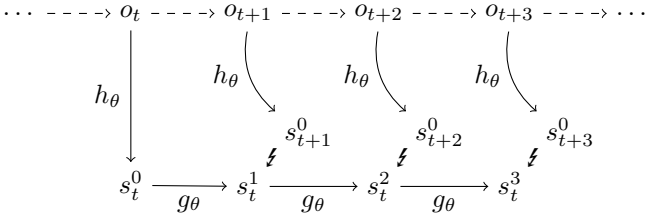

Say, as an example, we have two subsequent observations $o_t$ and $o_{t+1}$, between which action $a_t$ was performed. We can create their corresponding embedded state representations $s^0_t = h_\theta(o_1, ..., o_t)$ and $ s^0_{t+1} = h_\theta(o_1, ..., o_t, o_{t+1})$. By applying the dynamics function $g_\theta$ to $s_t^0$ as well as action $a_t$, we receive another state representation $s_t^1$ that is intuitively supposed to reflect the environment at timestep $t+1$, much like $s_{t+1}^0$. However, so long as both state representations allow for reward and value predictions, MuZero does not require them to match, or even be similar. This pattern persists with every iteration of $g_\theta$, as is apparent in Fig. \ref{fig:consistency_loss}. To clarify further, imagine a constructed but theoretically legitimate example in which state vector $s^0_{t+1}$ uses only the first half of its dimensions, whereas $s^1_t$ uses only the second half.

While the existence of multiple state formats may not be inherently destructive, and has been suggested to provide some benefits for the agent's understanding of the environment, we believe it can cause several problems. The most obvious is the need for the dynamics and prediction functions to learn how to use each of the different representations for the same estimates. If instead, $h_\theta$ and $g_\theta$ agree on a unified state format, this problem can be avoided. A more subtle but potentially more significant issue becomes apparent when we inspect MuZero's learning process. The functions are unrolled for $K$ timesteps across a given trajectory, and losses are computed for each timestep $k \in \{0, ..., K\}$. This means that $g_\theta$ and $f_\theta$ are trained on any state format that is produced by $g_\theta$ after up to $K$ iterations. For any further iterations, accuracy may degenerate. Depending on the search policy, it is not unlikely for the planning depth to become larger than $K$. In fact, the hyperparameters used for the original MuZero experiments have a small $K=5$ unroll depth for training, while at the same time using $50$ or even $800$ simulations for tree construction during planning, which can easily result in branches that are longer than $K$. By enforcing a consistent state representation, we may be able to mitigate the degeneration of performance for long planning branches.

We implement our change by continuously adjusting $\theta$ so that output $s^k_t$ of the dynamics function is close to output $s^0_{t+k}$ of the representation function. For example, we can perform gradient descent on the mean squared error between the state vectors. Mathematically, we express the consistency loss as $l^c(s^0_{t+k}, s^k_t)$, leading to an overall loss of
\begin{multline}
    l_t(\theta) = \sum^K_{k=0} \big(l^r(u_{t+k}, r_t^k) + l^v(z_{t+k}, v_t^k) \\
    + l^p(\pi_{t+k}, \policy_t^k) + l^c(s^0_{t+k}, s^k_t) + c||\theta||^2\big).
\end{multline}
Note that we treat $s^0_{t+k}$, the first parameter of $l^c$, as a constant, meaning the loss is not propagated directly to the representation function. Doing so would encourage $h_\theta$ and $g_\theta$ to always produce the same state outputs, regardless of the input. As a bonus, by only propagating the error through $s_t^k$ instead, the model is forced to maintain relevant information to predict subsequent states even in environments with sparse rewards, similar to the reconstruction loss from our previous proposal.

%% file: experiments/experiments.tex
\section{Experiments}
We will now showcase several experiments to test and ideally validate our hypothesis that the proposed methods improve the performance of the MuZero algorithm.

\input{experiments/environments.tex}
\input{experiments/setup.tex}

\input{experiments/results.tex}
\input{experiments/discussion.tex}

%% file: experiments/environments.tex
\subsection{Environments}
For our experiments, we choose environments provided by \textit{OpenAI Gym} \cite{gym}, which is a toolkit for developing and comparing reinforcement learning algorithms. It provides us with a simple and easy-to-use environment interface and a wide range of environments to develop general-purpose agents. More specifically, we choose two of the environments packaged with OpenAI Gym, namely \textit{CartPole-v1} and \textit{LunarLander-v2}.

A more meaningful benchmark would include significantly more complex environments, such as Chess, Go, and Atari games, as was done for the original MuZero agent. Furthermore, experiments with robot agents, like a robotic arm grasping for objects, could demonstrate real-world viability for the algorithms. Unfortunately, due to MuZero's high computational requirements, we are unable to properly test these environments with various hyperparameters and achieve sufficient statistical significance. The exploration of more demanding environments is therefore left to future work.

%% file: experiments/setup.tex
\subsection{Setup}
We adopt muzero-general \cite{muzero-general}, an open-source implementation of MuZero that is written in the Python programming language and uses PyTorch for automatic differentiation, as our baseline agent. It is heavily parallelized by employing a worker architecture, with each worker being a separate process that communicates with its peers through message passing. This allows for flexible scaling, even across multiple machines.

We modify the source code of muzero-general to include a reconstruction function and the additional loss terms. The readout of the replay buffer must also be tweaked to include not only the observation at timestep $t$, but the $K$ subsequent observations $o_{t+1}, ..., o_{t+K}$ as well, as they are critical for calculating our new loss values.

A variety of different weights are tested for each of the two loss terms in order to gauge their capability of improving performance, both individually and in a union. Furthermore, as a means of showcasing self-supervised learning for MuZero, we pretrain a hybrid agent, that is, an agent using both modifications at the same time, for $5000$ training steps using only the newly added loss terms instead of the full loss formula.

Table \ref{tab:hyperparameters} shows the hyperparameters used for all tested model configurations. Based on the muzero-general default parameters, only the number of training steps was reduced to be able to perform additional test runs with the newly freed resources. Note that we are unable to perform a comprehensive hyperparameter search due to computational limitations. Performance is measured by training an agent for a specific amount of training steps and, at various timesteps, sampling the total episode reward the agent can achieve.

\begin{table}
\caption{Hyperparameter selection for all tested agents. Parameters based on the current training step use the variable $t$. Only parameters marked with a $^*$ are different from the muzero-general defaults.}
    \centering
    \resizebox{\columnwidth}{!}{%
    \begin{tabular}{lrr}
        \hline
        & CartPole & LunarLander \\
        \hline

        Training steps & $10000^*$ & $30000^*$ \\
        Discount factor ($\gamma$) & 0.997 & 0.999 \\
        TD steps ($n$) & 50 & 30 \\
        Unroll steps ($K$) & 10 & 10 \\
        State dimensions & 8 & 10 \\
        MuZero Reanalyze & Enabled & Enabled \\

        \hline

        \multicolumn{3}{c}{Loss optimizer's parameters (Adam \cite{kingma2014adam})}\\
        \hline
        Learning rate ($\beta$) & $0.02 \times 0.9^{t \times 0.001}$ & 0.005 \\
        Value loss weight & 1.0 & 1.0 \\
        L2 reg. weight & $10^{-4}$ & $10^{-4}$ \\

        \hline

        \multicolumn{3}{c}{Replay parameters}\\
        \hline
        Replay buffer size & 500 & 2000 \\
        Prioritization exp. & 0.5 & 0.5 \\
        Batch size & 128 & 64 \\

        \hline

        \multicolumn{3}{c}{MCTS parameters} \\
        \hline
         Simulations & 50 & 50 \\
        Dirichlet $\alpha$ & 0.25 & 0.25 \\
        Exploration factor & 0.25 & 0.25 \\
        pUCT $c_1$ & 1.25 & 1.25 \\
        pUCT $c_2$ & 19652 & 19652 \\
        Temperature ($T$) & \makecell{
            1.0 if $t<5000$, \\ 0.5 if $5000 \leq t < 7500$, \\ 0.25 if $t \geq 7500$
        } & 0.35 \\

        \hline
    \end{tabular}
    }
    \label{tab:hyperparameters}
\end{table}

%% file: experiments/results.tex
\subsection{Results}
We show a comparison of the performance of agents with different weights applied to each loss term proposed in this paper. For our notation, we use $l^g$ and $l^c$ for the reconstruction function and consistency loss modification respectively. With $\frac{1}{2}l^g$ and $\frac{1}{2}l^c$ we denote that the default loss weight of $1$ for each term has been changed to $\frac{1}{2}$. Finally, we write a plus sign to indicate the combination of both modifications.

The results in Fig. \ref{fig:reconstruction_results} show an increase in performance when adding the reconstruction function together with its associated loss term to the MuZero algorithm on all testing environments. Weighting the reconstruction loss term with $\frac{1}{2}$ only has a minor improvement on the learning process. Note that, in the LunarLander-v2 environment, a penalty reward of $-100$ is given to an agent for crashing the lander. The default MuZero agent was barely able to exceed this threshold, whereas the reconstruction agent achieved positive total rewards.

The agent with the consistency loss term matched or only very slightly exceeded the performance of MuZero in the CartPole-v1 environment, as can be seen in Fig. \ref{fig:consistency_results} (left). However, in the LunarLander-v2 task, the modified agent significantly outperformed MuZero, being almost at the same level as the reconstruction agent. A loss weight of $1$ is also notably better than a loss weight of $\frac{1}{2}$.

\begin{figure}[t]
    \centering
    \begin{tikzpicture}[yscale=0.69, xscale=0.69,
                        define rgb/.code={\definecolor{mycolor}{RGB}{#1}},
                        rgb color/.style={define rgb={#1},mycolor}]
        \begin{axis}[
            title = CartPole-v1,
            axis lines = left,
            xlabel = Training steps,
            ylabel = Total reward,
            no markers,
            table/col sep = comma,
            legend cell align=left,
            legend pos=south east,
            legend style={draw=none},
            xmin=0,
            xmax=10000,
            grid=major,
            width=6cm,
            height=6cm,
        ]
            \addplot[black] table [
                x = training_step,
                y = reward,
            ] {results/default/CartPole-v1/lr0.0.csv};
            \addlegendentry{MuZero};

            \addplot[rgb color={128, 128, 255}] table [
                x = training_step,
                y = reward,
            ] {results/reconstruction/CartPole-v1/lr1.0.csv};
            \addlegendentry{$l^g$};

            \addplot[rgb color={0, 128, 128}] table [
                x = training_step,
                y = reward,
            ] {results/hybrid/CartPole-v1/lr0.5.csv};
            \addlegendentry{$\frac{1}{2}l^g + \frac{1}{2}l^c$};

            \addplot[rgb color={0, 200, 200}] table [
                x = training_step,
                y = reward,
            ] {results/hybrid/CartPole-v1/lr1.0.csv};
            \addlegendentry{$l^g + l^c$};
        \end{axis}
    \end{tikzpicture}
    \begin{tikzpicture}[yscale=0.69, xscale=0.69,
                        define rgb/.code={\definecolor{mycolor}{RGB}{#1}},
                        rgb color/.style={define rgb={#1},mycolor}]
        \begin{axis}[
            title = LunarLander-v2,
            axis lines = left,
            xlabel = Training steps,
            ylabel = Total reward,
            no markers,
            table/col sep = comma,
            legend cell align=left,
            legend pos=south east,
            legend style={draw=none},
            xmin=0,
            xmax=30000,
            grid=major,
            width=6cm,
            height=6cm,
        ]
            \addplot[black] table [
                x = training_step,
                y = reward,
            ] {results/default/LunarLander-v2/lr0.0.csv};
            \addlegendentry{MuZero};

            \addplot[rgb color={128, 128, 255}] table [
                x = training_step,
                y = reward,
            ] {results/reconstruction/LunarLander-v2/lr1.0.csv};
            \addlegendentry{$l^g$};

            \addplot[rgb color={0, 128, 128}] table [
                x = training_step,
                y = reward,
            ] {results/hybrid/LunarLander-v2/lr0.5.csv};
            \addlegendentry{$\frac{1}{2}l^g + \frac{1}{2}l^c$};

            \addplot[rgb color={0, 200, 200}] table [
                x = training_step,
                y = reward,
            ] {results/hybrid/LunarLander-v2/lr1.0.csv};
            \addlegendentry{$l^g + l^c$};
        \end{axis}
    \end{tikzpicture}
    \caption{Total episode reward comparison of agents using both the reconstruction function loss ($l^g$) as well as the consistency loss term ($l^c$) simultaneously in the CartPole-v1 and LunarLander-v2 environments, averaged across 32 and 25 runs, respectively.}
    \label{fig:hybrid_results}
\end{figure}
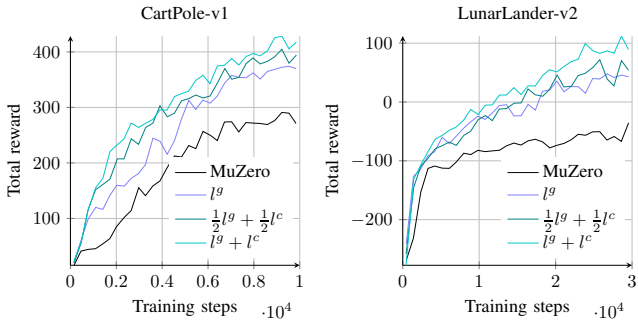

An agent using both loss terms simultaneously outperforms MuZero (cf.~Fig. \ref{fig:hybrid_results}), and even scores marginally better than the reconstruction loss agent, in all environments tested.

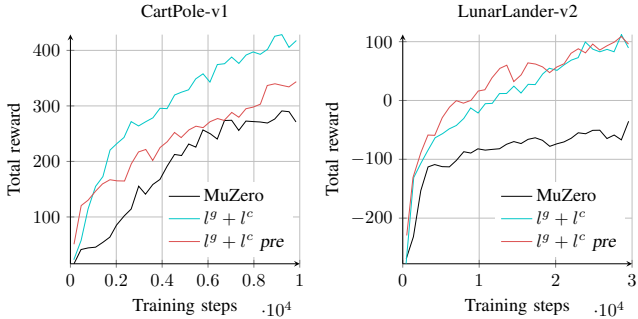
\begin{figure}
    \centering
    \begin{tikzpicture}[yscale=0.69, xscale=0.69,
                        define rgb/.code={\definecolor{mycolor}{RGB}{#1}},
                        rgb color/.style={define rgb={#1},mycolor}]
        \begin{axis}[
            title = CartPole-v1,
            axis lines = left,
            xlabel = Training steps,
            ylabel = Total reward,
            no markers,
            table/col sep = comma,
            legend cell align=left,
            legend pos=south east,
            legend style={draw=none},
            xmin=0,
            xmax=10000,
            grid=major,
            width=6cm,
            height=6cm,
        ]
            \addplot[black] table [
                x = training_step,
                y = reward,
            ] {results/default/CartPole-v1/lr0.0.csv};
            \addlegendentry{MuZero};

            \addplot[rgb color={0, 200, 200}] table [
                x = training_step,
                y = reward,
            ] {results/hybrid/CartPole-v1/lr1.0.csv};
            \addlegendentry{$l^g + l^c$};

            \addplot[rgb color={220, 80, 80}] table [
                x = training_step,
                y = reward,
            ] {results/pretrained/CartPole-v1/lr1.0.csv};
            \addlegendentry{$l^g + l^c$ {\sl pre}};
        \end{axis}
    \end{tikzpicture}
    \begin{tikzpicture}[yscale=0.69, xscale=0.69,
                        define rgb/.code={\definecolor{mycolor}{RGB}{#1}},
                        rgb color/.style={define rgb={#1},mycolor}]
        \begin{axis}[
            title = LunarLander-v2,
            axis lines = left,
            xlabel = Training steps,
            ylabel = Total reward,
            no markers,
            table/col sep = comma,
            legend cell align=left,
            legend pos=south east,
            legend style={draw=none},
            xmin=0,
            xmax=30000,
            grid=major,
            width=6cm,
            height=6cm,
        ]
            \addplot[black] table [
                x = training_step,
                y = reward,
            ] {results/default/LunarLander-v2/lr0.0.csv};
            \addlegendentry{MuZero};

            \addplot[rgb color={0, 200, 200}] table [
                x = training_step,
                y = reward,
            ] {results/hybrid/LunarLander-v2/lr1.0.csv};
            \addlegendentry{$l^g + l^c$};

            \addplot[rgb color={220, 80, 80}] table [
                x = training_step,
                y = reward,
            ] {results/pretrained/LunarLander-v2/lr1.0.csv};
            \addlegendentry{$l^g + l^c$ {\sl pre}};
        \end{axis}
    \end{tikzpicture}
    \caption{Total episode reward comparison of agents using both the reconstruction function loss ($l^g$) as well as the consistency loss term ($l^c$) simultaneously as a non-pretrained and pretrained (denoted {\sl pre}) variant in the CartPole-v1 and LunarLander-v2 environments averaged across 32 and 25 runs, respectively.}
    \label{fig:pretrained_results}
\end{figure}

When using self-supervised pretraining (see Fig. \ref{fig:pretrained_results}), training progresses very rapidly as soon as the goal is introduced. In the LunarLander-v2 environment, a mean total reward of $0$ is reached in roughly half the amount of training steps that are required by the non-pretrained agent. However, at later stages of training, the advantage fades, and, in the case of CartPole-v1, agents using self-supervised pretraining perform significantly worse than agents starting with randomly initialized networks.

The trained agents are compared in Table \ref{tab:results_table}. Training took place on NVIDIA GeForce GTX 1080 and NVIDIA GeForce RTX 2080 Ti GPUs. Each experiment required roughly two to three days to complete.

\begin{figure}
    \centering
    \begin{tikzpicture}[yscale=0.69, xscale=0.69,
                        define rgb/.code={\definecolor{mycolor}{RGB}{#1}},
                        rgb color/.style={define rgb={#1},mycolor}]
        \begin{axis}[
            title = CartPole-v1,
            axis lines = left,
            xlabel = Training steps,
            ylabel = Total reward,
            no markers,
            table/col sep = comma,
            legend cell align=left,
            legend pos=south east,
            legend style={draw=none},
            xmin=0,
            xmax=10000,
            grid=major,
            width=6cm,
            height=6cm,
        ]
            \addplot[black] table [
                x = training_step,
                y = reward,
            ] {results/default/CartPole-v1/lr0.0.csv};
            \addlegendentry{MuZero};

            \addplot[rgb color={32, 32, 180}] table [
                x = training_step,
                y = reward,
            ] {results/reconstruction/CartPole-v1/lr0.5.csv};
            \addlegendentry{$\frac{1}{2}l^g$};

            \addplot[rgb color={128, 128, 255}] table [
                x = training_step,
                y = reward,
            ] {results/reconstruction/CartPole-v1/lr1.0.csv};
            \addlegendentry{$l^g$};
        \end{axis}
    \end{tikzpicture}
    \begin{tikzpicture}[yscale=0.69, xscale=0.69,
                        define rgb/.code={\definecolor{mycolor}{RGB}{#1}},
                        rgb color/.style={define rgb={#1},mycolor}]
        \begin{axis}[
            title = LunarLander-v2,
            axis lines = left,
            xlabel = Training steps,
            ylabel = Total reward,
            no markers,
            table/col sep = comma,
            legend cell align=left,
            legend pos=south east,
            legend style={draw=none},
            xmin=0,
            xmax=30000,
            grid=major,
            width=6cm,
            height=6cm,
        ]
            \addplot[black] table [
                x = training_step,
                y = reward,
            ] {results/default/LunarLander-v2/lr0.0.csv};
            \addlegendentry{MuZero};

            \addplot[rgb color={32, 32, 180}] table [
                x = training_step,
                y = reward,
            ] {results/reconstruction/LunarLander-v2/lr0.5.csv};
            \addlegendentry{$\frac{1}{2}l^g$};

            \addplot[rgb color={128, 128, 255}] table [
                x = training_step,
                y = reward,
            ] {results/reconstruction/LunarLander-v2/lr1.0.csv};
            \addlegendentry{$l^g$};
        \end{axis}
    \end{tikzpicture}
    \caption{Total episode reward comparison of agents using the reconstruction loss term ($l^g$) and the default MuZero agent in the CartPole-v1 and LunarLander-v2 environments, averaged across 32 and 25 runs, respectively.}
    \label{fig:reconstruction_results}
\end{figure}
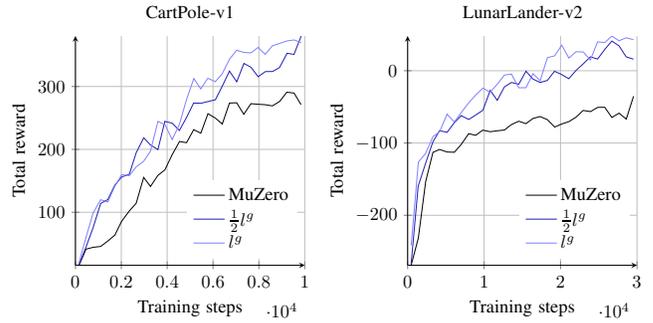

\begin{figure}
    \centering
    \begin{tikzpicture}[yscale=0.69, xscale=0.69,
                        define rgb/.code={\definecolor{mycolor}{RGB}{#1}},
                        rgb color/.style={define rgb={#1},mycolor}]
        \begin{axis}[
            title = CartPole-v1,
            axis lines = left,
            xlabel = Training steps,
            ylabel = Total reward,
            no markers,
            table/col sep = comma,
            legend cell align=left,
            legend pos=south east,
            legend style={draw=none},
            xmin=0,
            xmax=10000,
            grid=major,
            width=6cm,
            height=6cm,
        ]
            \addplot[black] table [
                x = training_step,
                y = reward,
            ] {results/default/CartPole-v1/lr0.0.csv};
            \addlegendentry{MuZero};

            \addplot[rgb color={32, 128, 32}] table [
                x = training_step,
                y = reward,
            ] {results/consistency/CartPole-v1/lr0.5.csv};
            \addlegendentry{$\frac{1}{2}l^c$};

            \addplot[rgb color={0, 255, 0}] table [
                x = training_step,
                y = reward,
            ] {results/consistency/CartPole-v1/lr1.0.csv};
            \addlegendentry{$l^c$};
        \end{axis}
    \end{tikzpicture}
    \begin{tikzpicture}[yscale=0.69, xscale=0.69,
                        define rgb/.code={\definecolor{mycolor}{RGB}{#1}},
                        rgb color/.style={define rgb={#1},mycolor}]
        \begin{axis}[
            title = LunarLander-v2,
            axis lines = left,
            xlabel = Training steps,
            ylabel = Total reward,
            no markers,
            table/col sep = comma,
            legend cell align=left,
            legend pos=south east,
            legend style={draw=none},
            xmin=0,
            xmax=30000,
            grid=major,
            width=6cm,
            height=6cm,
        ]
            \addplot[black] table [
                x = training_step,
                y = reward,
            ] {results/default/LunarLander-v2/lr0.0.csv};
            \addlegendentry{MuZero};

            \addplot[rgb color={32, 128, 32}] table [
                x = training_step,
                y = reward,
            ] {results/consistency/LunarLander-v2/lr0.5.csv};
            \addlegendentry{$\frac{1}{2}l^c$};

            \addplot[rgb color={0, 255, 0}] table [
                x = training_step,
                y = reward,
            ] {results/consistency/LunarLander-v2/lr1.0.csv};
            \addlegendentry{$l^c$};
        \end{axis}
    \end{tikzpicture}
    \caption{Total episode reward comparison of agents using the consistency loss ($l^c$) and the default MuZero agent in the CartPole-v1 and LunarLander-v2 environments, averaged across 32 and 25 runs, respectively.}
    \label{fig:consistency_results}
\end{figure}

\begin{table}
\caption{Comparison of the default MuZero algorithm and the modifications described in this paper on the CartPole-v1 and LunarLander-v2 environments. The terms $l^g$ and $l^c$ stand for the addition of a reconstruction or consistency loss, respectively, {\sl pre} for pretrained. The results show the mean and standard deviation of the total reward for the final $500$ training steps across all test runs.}
    \centering
    \begin{tabular}{lcc}
        \hline
        & CartPole-v1 & LunarLander-v2 \\
        \hline
        MuZero & $281.42 \pm 162.48$ & $-34.86 \pm 92.87$ \\
        $l^g$ & $375.85 \pm 149.38$ & $42.67 \pm 132.59$\\
        $l^c$ & $296.45 \pm 174.55$ & $32.17 \pm 145.90$ \\
        $l^g + l^c$ & $\mathbf{410.59 \pm 130.71}$ & $100.46 \pm 123.82$ \\
        $l^g + l^c$ {\sl pre} & $335.72 \pm 162.49$ & $\mathbf{104.99 \pm 116.67}$ \\
        \hline
    \end{tabular}
    \label{tab:results_table}
\end{table}

%% file: experiments/discussion.tex
\subsection{Discussion}
The results show that all modified agents we tested exceed the performance of the unmodified MuZero agent on all environments, some, especially the agents including both proposed changes simultaneously, by a large amount. The reconstruction function seems to be the more influential one of the two changes, given that it achieved a bigger performance increase than the consistency loss, and comes very close to even the hybrid agent. Even so, the consistency loss has the advantage of being simpler in its implementation by not requiring a fourth function, potentially made up of complex neural networks requiring domain expertise for their parameterization, to be added.

Agents that were pretrained in a self-supervised fashion ahead of time boosted initial training but were unable to outmatch their non-pretrained counterparts. This is to be expected. While self-supervised pretraining is meant to accelerate learning, there is no argument as to why training with partial information should yield higher maximum scores in the long run. In CartPole-v1, the pretrained agent even performed significantly worse than the non-pretrained version after roughly $1000$ training steps. This is likely due to the parameters and internal state representations of the model prematurely converging, to some extent, to a local minimum, which made reward and value estimations more difficult after the goal was introduced. Use of L2 regularization during pretraining may prevent this unwanted convergence.

While the consistency loss resulted in a performance increase, the experiments were not designed to confirm that state representations did indeed become more consistent. Future work should also validate our hypothesis stating that the consistency loss may lessen the accuracy falloff when planning further than $K$ timesteps into the future, with $K$ being the number of unrolled steps during training.

Due to computational limitations we did not simulate computationally complex environments such as Go, Atari games, or vision-driven robotics scenarios. Furthermore, an extensive hyperparameter search regarding the optimal combination of loss weights is yet to be performed, particularly because our best-performing agents were the ones with the highest weight settings. We leave these topics to future work.

%% file: conclusion/conclusion.tex
\section{Conclusion}
We have proposed changes to the MuZero Algorithm consisting of two new loss terms to the overall loss function. One of these requires an auxiliary neural network to be introduced and allows unsupervised pretraining of MuZero agents. Experiments on simple OpenAI Gym environments have shown that they significantly increase the model's performance, especially when used in combination.

The full source code used for the experiments is available on GitHub at \url{https://github.com/pikaju/muzero-g}.